\documentclass[conference]{IEEEtran}
\IEEEoverridecommandlockouts
\usepackage{cite}
\usepackage{amsmath,amssymb,amsfonts}
\usepackage{algorithmic}
\usepackage{graphicx}
\usepackage{textcomp}
\usepackage{xcolor, colortbl}
\usepackage{soul}

\definecolor{blue_highlight}{RGB}{173, 216, 230}  
\definecolor{green_highlight}{RGB}{144, 238, 144} 

\usepackage[utf8]{inputenc}
\usepackage[T1]{fontenc} 
\usepackage{url}

\def\BibTeX{{\rm B\kern-.05em{\sc i\kern-.025em b}\kern-.08em
    T\kern-.1667em\lower.7ex\hbox{E}\kern-.125emX}}
\begin{document}

\title{Vision-based Perception for Autonomous Vehicles in Obstacle Avoidance Scenarios\\
}

\author{\IEEEauthorblockN{Van-Hoang-Anh Phan}
\IEEEauthorblockA{\textit{ISLab, HCMUTE} \\
Ho Chi Minh City, Vietnam \\
21151070@student.hcmute.edu.vn}
\and
\IEEEauthorblockN{Chi-Tam Nguyen}
\IEEEauthorblockA{\textit{ISLab, HCMUTE} \\
Ho Chi Minh City, Vietnam \\
20133087@student.hcmute.edu.vn}
\and
\IEEEauthorblockN{Doan-Trung Au}
\IEEEauthorblockA{\textit{ISLab, HCMUTE} \\
Ho Chi Minh City, Vietnam \\
audoantrung123@gmail.com}
\and
\IEEEauthorblockN{Thanh-Danh Phan}
\IEEEauthorblockA{\textit{Dept. Intelligent Systems $\&$ Robotics} \\
\textit{Chungbuk National University}\\
Cheongju, South Korea\\
 danh1711@chungbuk.ac.kr}
\and
\IEEEauthorblockN{Minh-Thien Duong}
\IEEEauthorblockA{\textit{Dept. Automatic Control, HCMUTE} \\
Ho Chi Minh City, Vietnam \\
minhthien@hcmute.edu.vn}
\and
\IEEEauthorblockN{My-Ha Le}
\IEEEauthorblockA{\textit{Faculty of Electrical and Electronics Engineering} \\
\textit{HCMC University of Technology and Education}\\
Ho Chi Minh City, Vietnam \\
Corresponding author:\\
halm@hcmute.edu.vn}\\
}

\maketitle

\begin{abstract}
Obstacle avoidance is essential for ensuring the safety of autonomous vehicles. Accurate perception and motion planning are crucial to enabling vehicles to navigate complex environments while avoiding collisions. In this paper, we propose an efficient obstacle avoidance pipeline that leverages a camera-only perception module and a Frenet-Pure Pursuit-based planning strategy. By integrating advancements in computer vision, the system utilizes YOLOv11 for object detection and state-of-the-art monocular depth estimation models, such as Depth Anything V2, to estimate object distances. A comparative analysis of these models provides valuable insights into their accuracy, efficiency, and robustness in real-world conditions. The system is evaluated in diverse scenarios on a university campus, demonstrating its effectiveness in handling various obstacles and enhancing autonomous navigation. The video presenting the results of the obstacle avoidance experiments is available at: \url{https://www.youtube.com/watch?v=FoXiO5S_tA8}
\end{abstract}


\section{Introduction}
Autonomous vehicles have become a major topic in the research community \cite{duong2018navigating, e2ecar24, vancar20}, particularly in the development of safety systems for self-driving cars \cite{nguyencrossing24, WangSafe20, CarlosARS13}. Among these, obstacle avoidance is a critical area of interest. The objective of this task is to design an algorithm that enables smooth navigation while ensuring collision avoidance. Additionally, the algorithm must be efficient and compatible with real-world hardware constraints. 

The obstacle avoidance process relies on two key processes: perception and planning \cite{CarlosARS13, LaghmaraIV19}. Perception is responsible for collecting and interpreting data from the surrounding environment, such as detecting obstacles and estimating distances. Some existing methods \cite{CarlosARS13, Zhao3DLIDAR20} utilize 3D LiDAR data to determine the position of obstacles. The authors of \cite{NguyenIWIS22} proposed a fusion of a camera and 2D LiDAR to identify drivable areas and determine avoidance directions. In this approach, the 2D LiDAR monitors scene information on the right side, which lies outside the camera’s field of view. However, due to cost constraints and system simplicity, we use only a single camera for obstacle detection and distance estimation.

The safe and reliable planning of obstacle-avoidance routes is a critical aspect of autonomous vehicle technology. Several research efforts have explored different methodologies to address this challenge \cite{Weng2024, Yoon2025, Yijian22apf}. Genetic algorithms have been employed as a path planning technique for autonomous vehicles. Most representatively, Weng et al. \cite{Weng2024} particularly emphasizing the integration of the Frenet-Serret coordinate system and Sequential Quadratic Programming (SQP) to enhance adaptability and optimize trajectories. Yoon et al. \cite{Yoon2025} proposed an alternative approach that combines a neural network for path proposal with sampling-based path planning in the Frenet coordinate system. In addition, traditional methods such as the Artificial Potential Field (APF) have also been utilized for obstacle avoidance \cite{Yijian22apf}. 

 In our system, the Frenet coordinate system is employed for path planning in structured road environments, providing an efficient representation for trajectory generation. More specifically, the Pure Pursuit algorithm is utilized for accurate path tracking to ensure precise and stable vehicle control. By integrating these approaches, our system facilitates smooth and stable vehicle motion while enhancing obstacle avoidance capabilities.

\begin{figure*}[t]
\centering
\includegraphics[width=1\textwidth]{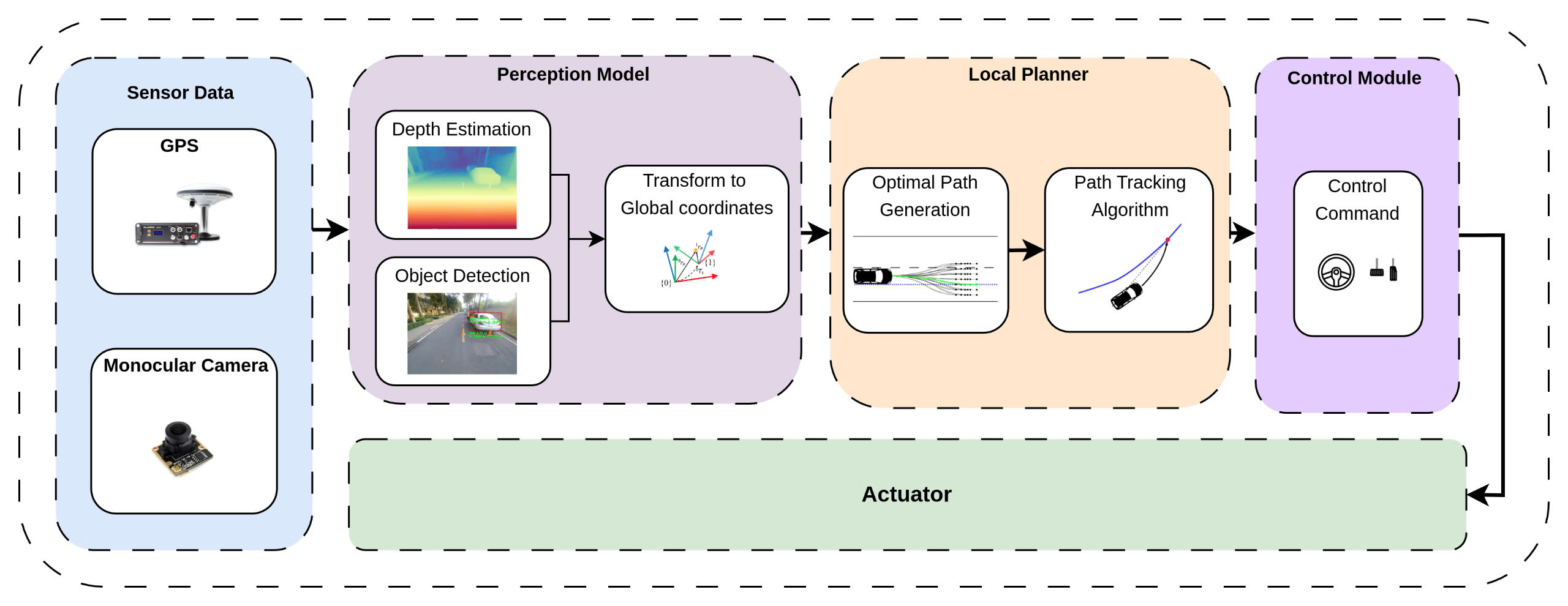}
\caption{The block diagram of the proposed method.}
\label{fig1}
\end{figure*}

Based on the analysis above, this paper proposes an efficient perception and planning system for autonomous obstacle avoidance, utilizing a camera-only perception module and a Frenet-Pure Pursuit-based motion planning strategy. We developed an obstacle avoidance system for autonomous vehicles that can operate in real-world environments. The perception module relies solely on a monocular camera to capture environmental information. Obstacles are detected using a You Only Look Once (YOLO) object detection model, where we evaluate and compare state-of-the-art versions, including YOLOv9 \cite{wang2024yolov9}, YOLOv10 \cite{wang2024yolov10}, and YOLOv11 \cite{yolo11_ultralytics}, to select the most suitable model for obstacle detection. Their distances are estimated using Depth Anything V2 \cite{yang2024depthv2}, a monocular depth estimation foundation model. Furthermore, we conduct experiments and analyze various monocular depth estimation foundation models in real-world scenarios to determine the most effective one. The planning module integrates Frenet and Pure Pursuit methods for optimal path generation and path tracking, as illustrated in Fig. \ref{fig1}.

In summary, this paper presents the following key contributions:

- Proposes an efficient obstacle avoidance pipeline using a camera-only perception module and a Frenet-Pure Pursuit-based planning strategy.

- Evaluates and compares state-of-the-art YOLO models for obstacle detection and selects the most suitable model. Additionally, analyzes monocular depth estimation models in real-world environments to identify the most accurate and reliable model for obstacle distance estimation.

- Conducts extensive real-world experiments across various scenarios to analyze the effectiveness of the proposed system in obstacle avoidance and path planning.

\section{METHODOLOGY}

\subsection{Perception Module}
The perception module is crucial for the navigation of autonomous vehicles as it enables the vehicle to understand its surroundings. It integrates data from an RTK GPS and a camera to detect and localize obstacles. The system comprises three major components: object detection, depth estimation, and coordinate transformation. Among them, object detection takes on identifying obstacles, depth estimation model is used to calculate their distances, and coordinate transformation algorithm aims at converting image coordinates into the global coordinate system.

\subsubsection{Obstacle detection} In this work, we first evaluate and compare multiple state-of-the-art YOLO models to determine the most suitable one for real-world deployment. After extensive evaluation based on criteria such as accuracy, computational efficiency, and real-time performance, YOLOv11 \cite{yolo11_ultralytics} is selected for its significant detection accuracy and promising real-time performance in identifying and classifying objects in diverse environments. The detected obstacles are assigned bounding boxes, which provide the initial spatial information used for further depth estimation in the subsequent stages.

\subsubsection{Monocular Depth Estimation} A depth map is vital for understanding the 3D structure of the environment and ensuring accurate distance estimation between the vehicle and obstacles. To this end, we employ Depth Anything V2 \cite{yang2024depthv2}, a high-performance monocular depth estimation model trained on extensive datasets. In our evaluation, we compare the performance of several state-of-the-art depth estimation models, considering factors such as the discrepancy between estimated and actual distances, as well as computational efficiency. By integrating the depth map from Depth Anything V2 \cite{yang2024depthv2} with the bounding box coordinates from YOLOv11 \cite{yolo11_ultralytics}, our system accurately determines the real-world positions of detected obstacles. This fusion significantly enhances situational awareness, enabling the vehicle to navigate complex environments more precisely while avoiding obstacles.

\subsubsection{Coordinate Transformation} To enable local path planning and vehicle control, we establish coordinate transformations between the vehicle frame and the global Universal Transverse Mercator (UTM) coordinate system. The obstacle localization process involves three key coordinate transformations, as detailed below.

- \textit{Image to Camera Coordinates.} For each detected obstacle with bounding box coordinates $(x_i, y_i)$, we first compute the median depth value $d$ within the bounding box to mitigate depth estimation noise. The transformation from pixel coordinates $(x_i, y_i)$ to 3D camera coordinates $P_{cam}=[x_{cam}, y_{cam}, z_{cam}]^{T}$ is derived using the camera intrinsic matrix $K$:

\begin{equation}
P_{cam} = d \cdot K^{-1} \begin{bmatrix} x_i & y_i & 1\end{bmatrix}^T
\label{euq:2}
\end{equation}
where \( K = \begin{bmatrix} f_x & 0 & c_x \\ 0 & f_y & c_y \\ 0 & 0 & 1 \end{bmatrix} \) contains the focal lengths \( (f_x, f_y) \) and principal point \( (c_x, c_y) \).

- \textit{Camera to Vehicle Coordinates.} The camera-to-vehicle transformation accounts for the camera’s mounting position and orientation. Let $T^{veh}_{cam} \in \mathbb{R}^{4\times4}$ denote the homogeneous transformation matrix:

\begin{equation}
T^{veh}_{cam} = \begin{bmatrix} R^{veh}_{cam} & t^{veh}_{cam} \\ 0 & 1 \end{bmatrix}
\end{equation}
where:
\begin{itemize}
    \item $R^{veh}_{cam}$ denotes the rotation matrix defined by the camera’s downward tilt angle $\theta$ about the x-axis.
    \item $t = \begin{bmatrix} t_x & t_y & h\end{bmatrix}^T $ represents the translation vector, with $h$ as the camera height and $(t_x, t_y)$ as the lateral offset from the vehicle’s center.
    \item The extrinsic camera parameters are determined through the setup shown in Fig. \ref{fig_coordinate_sys_obstacle_avoidance}, including the camera height $h$, the tilt angle $\theta$ about the x-axis, and the lateral offset $(t_x, t_y)$. In this setup, $t_x = 0$ is defined.
\end{itemize}

The vehicle-frame coordinates $P_{veh} = \begin{bmatrix} x_{veh} & y_{veh} & z_{veh}\end{bmatrix}$ are computed as:

\begin{equation}
P_{veh} = T^{veh}_{cam} P_{cam}
\label{euq:2i}
\end{equation}

- \textit{Vehicle to Global Coordinates.} The final transformation to global UTM coordinates $P_{veh} = \begin{bmatrix} x_{global} & y_{global} & z_{global}\end{bmatrix}^T$ incorporates the vehicle’s GPS position $(x_{g}, y_{g}, x_{g})$ and heading $\psi$ from the compass. The transformation is defined as:

\begin{equation}
P_{global} = T^{global}_{veh} P_{veh} = \begin{bmatrix} R^{global}_{veh} & t^{global}_{veh} \\ 0 & 1 \end{bmatrix} P_{veh}
\end{equation}
where:
\begin{itemize}
    \item $R^{global}_{veh}$ expresses the rotation matrix aligned with the vehicle’s heading.
    \item The translation vector $t^{global}_{veh}$ accounts for the camera's mounting position relative to the vehicle and the GPS antenna's mounting point.
\end{itemize}

\subsection{Local Planner}

\subsubsection{Optimal Path Generation} Frenet Optimal Trajectory (FOT) method \cite{WerlingFrenet10} enables autonomous vehicles to generate collision-free paths by decomposing motion into longitudinal and lateral components. Lateral motion is modeled using quintic polynomials for smooth transitions, while longitudinal motion employs quartic polynomials to maintain continuity. Candidate trajectories with varying lateral offsets and speeds are evaluated using a cost function that penalizes jerk, travel time, lane deviation, and speed errors, ensuring an optimal balance of smoothness, safety, and feasibility. In our project, the FOT method is utilized for path tracking and obstacle avoidance, with refinements made by adjusting cost function weights and tuning parameters to better align with vehicle dynamics. Additionally, trajectory updates are synchronized with the perception module, integrating RTK GPS and camera data for real-time responsiveness.

\subsubsection{Path Tracking} Pure Pursuit controllers are essential for accurate path tracking in autonomous vehicles. Model Predictive Control (MPC) offers high precision by optimizing control inputs over a prediction horizon. However, its computational demands introduce latency that limits real-time applicability. The Stanley Controller \cite{stanleycontroller}, a geometry-based method, minimizes cross-track error by adjusting steering based on the vehicle’s front-wheel position but struggles with abrupt path changes and external disturbances. In contrast, the Pure Pursuit Controller (PPC) determines steering angles using a look-ahead point, balancing robustness and efficiency. We selected PPC due to its stability and low computational cost. Moreover, it can dynamically adjust the look-ahead distance based on speed and road curvature. In our implementation, PPC tracked the optimal path generated by the Frenet Optimal Trajectory (FOT) method, ensuring smooth and accurate vehicle motion, with a lateral error below 0.2 m at 10 km/h, enabling reliable tracking in campus environments.

\subsection{Control Command and Actuation}
After determining the optimal trajectory and selecting the path-tracking controller, control commands for steering, acceleration, and braking are generated. These commands are sent to the vehicle’s actuators, which adjust the steering, throttle, and brakes to follow the desired path. Control signals are updated in real-time based on sensor feedback, allowing the vehicle to adapt to dynamic road conditions and obstacles.

\section{EXPERIMENTAL RESULTS}

\subsection{Autonomous Vehicle Hardware System}
\begin{figure}[t]
\centering
\includegraphics[width=1\columnwidth]{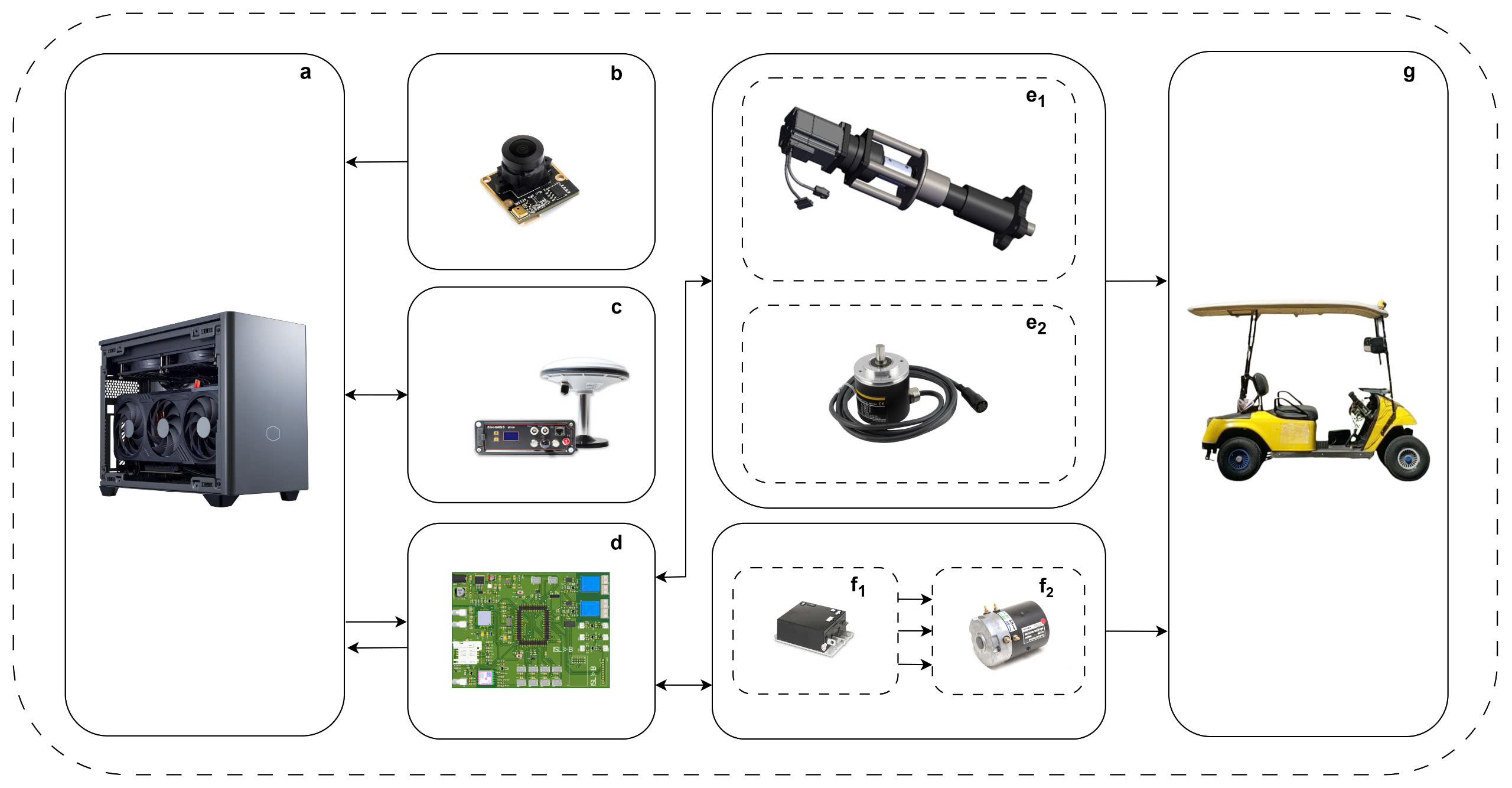}
\caption{Autonomous vehicle hardware system. (a) Central computer, (b) Camera, (c) RTK GPS, (d) PCB board,  ($e_1$) Steering motor, ($e_2$) Absolute encoder, ($f_1$) Motor controller, ($f_2$) DC shunt motor, (g) Autonomous Electric Golf Cart}
\label{hardware_systems}
\end{figure}

\begin{figure}[t]
\centering
\includegraphics[width=1\columnwidth]{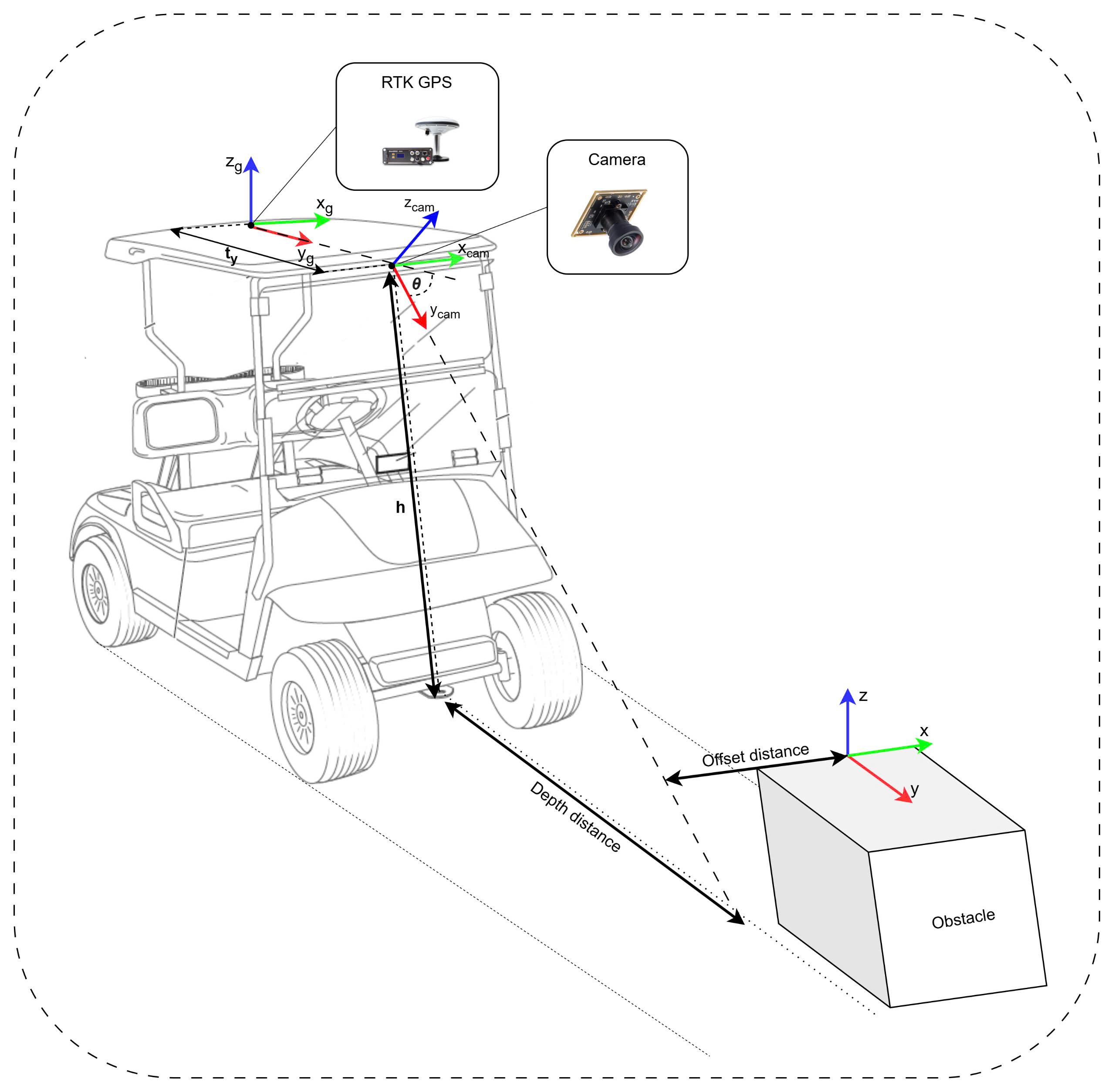}
\caption{Coordinate system and distance measurement for obstacle avoidance in autonomous vehicles}
\label{fig_coordinate_sys_obstacle_avoidance}
\end{figure}

The self-driving car is built on a modified EZGO electric golf cart, as shown in Fig. \ref{hardware_systems}. It is equipped with a central computer, a camera, an RTK GPS, a PCB board, a steering motor, an absolute encoder, a motor controller, and a DC shunt motor. The central computer features a 12th Gen Intel® Core™ i7-12700H processor (2.70 GHz), an NVIDIA GeForce GTX 3050 Ti GPU, and 8GB of RAM.

\subsection{Evaluation of Perception Performance in Real-World Environments}

\subsubsection{Obstacle detection}

\begin{table}[t]
\centering
\caption{Performance Evaluation of YOLO Models in Obstacle Detection}
\label{tab2}
\setlength{\tabcolsep}{3.5pt}
\def\arraystretch{2.5}%
\begin{tabular}{c|c| c |c| c| c }
\hline
Model & Params & Precision & Recall & mAP50 & FPS \\ \hline 
YOLOv9t & 1.9 & 97.7 & 99.5 & 99.4 & 60 \\
YOLOv10n & 2.7 & 97.7 & 97.5 & 99.3 & 80 \\
YOLOv11n & 2.6 & 97.3 & 98.6 & 99.3 & 84 \\
\hline
\end{tabular}
\end{table}

To evaluate the performance of obstacle detection in a real-world environment, we annotated 3,000 samples, splitting them into training, validation, and test sets with a ratio of 70:15:15. The dataset was collected from our environment (HCMUTE campus) and used for training and evaluation with state-of-the-art YOLO models, including YOLOv9 \cite{wang2024yolov9}, YOLOv10 \cite{wang2024yolov10}, and YOLOv11 \cite{yolo11_ultralytics}. In this context, obstacles primarily refer to static vehicles typically encountered in a university campus setting—such as cars, motorcycles, bicycles, trucks, delivery vans, and electric carts—which often obstruct the path of autonomous vehicles. Table \ref{tab2} presents the results, indicating minimal differences across key metrics such as the number of parameters, precision, recall, and mAP50. However, a significant variation is observed in FPS, with YOLOv9 \cite{wang2024yolov9} achieving 60 FPS, YOLOv10 \cite{wang2024yolov10} reaching 80 FPS, and YOLOv11 \cite{yolo11_ultralytics} performing best at 84 FPS. Given this difference in processing speed, we choose YOLOv11 \cite{yolo11_ultralytics} as the optimal model for real-time applications.

\subsubsection{Monocular depth estimation}

\begin{table}[t]
\caption{Comparison of depth estimation methods with ground truth using the \textit{depth distance} metric. Each value represents the absolute error in meters. The lowest errors are marked in \textcolor{blue}{blue}, and the second lowest errors are marked in \textcolor{red}{red}. The metric \textit{depth distance} is defined as shown in Fig.\ref{fig_coordinate_sys_obstacle_avoidance}.}

\label{tab:depth_comparison}
\begin{center}
\resizebox{\linewidth}{!}{%
\begin{tabular}{c|c|c|c}
\hline
\textbf{Ground Truth (m)} & \textbf{Depth Anything V2} & \textbf{MiDaS} & \textbf{Monodepth2} \\ \hline
\textbf{3}  & \textcolor{red}{0.148}  & \textcolor{blue}{0.102}   & 0.222  \\
\textbf{5}  & \textcolor{blue}{0.047}  & \textcolor{red}{0.392}  & 0.516  \\
\textbf{8}  & \textcolor{blue}{0.080}  & \textcolor{red}{0.156}  & 3.919  \\
\textbf{15} & \textcolor{blue}{0.366}  & 5.353  & \textcolor{red}{3.059}  \\ \hline
\end{tabular}
}
\end{center}
\end{table}

\begin{table}[t]
\caption{Comparison of depth estimation methods with ground truth using the \textit{offset distance} metric. Each value represents the offset in meters. The lowest errors are highlighted in \textcolor{blue}{blue}, and the second lowest errors are highlighted in \textcolor{red}{red}. The metric \textit{offset distance}  is defined as shown in Fig. \ref{fig_coordinate_sys_obstacle_avoidance}.}

\begin{center}
\resizebox{\linewidth}{!}{%
\setlength{\tabcolsep}{8pt} 
\renewcommand{\arraystretch}{1.3} 
\begin{tabular}{c|c|c|c}
\hline
\textbf{Ground Truth (m)} & \textbf{Depth Anything V2} & \textbf{MiDaS} & \textbf{Monodepth2} \\ \hline
\textbf{0}  & 0.073  & \textcolor{blue}{0.011}  & \textcolor{red}{0.058}  \\
\textbf{1}  & \textcolor{blue}{0.045}  & \textcolor{red}{0.202}  & 0.282  \\
\textbf{2}  & \textcolor{blue}{0.047}  & \textcolor{red}{0.114}  & 0.126  \\
\textbf{3}  & \textcolor{blue}{0.051}  & \textcolor{red}{0.104}  & 0.330  \\ \hline
\end{tabular}
}
\end{center}
\label{tab:depth_comparison_2}
\end{table}

\begin{table}[t]
\caption{Comparison of Computational Efficiency in Depth Estimation Models}
\begin{center}
\setlength{\tabcolsep}{8pt}
\centering
\def\arraystretch{1.3}%
\begin{tabular}{c|c|c|c}
\hline
\textbf{Model} & \textbf{Params (M)} & \textbf{FLOPs (G)} & \textbf{FPS} \\ \hline
Depth Anything V2    & 24.18  & 41.30  & 20  \\  
MiDaS          & 344  & 526.40  & 11  \\  
Monodepth2     & 14.33  & 8.04  & 31 \\ \hline
\end{tabular}\label{tab:params_flops_fps}
\end{center}
\end{table}

\begin{figure}[t]
\centering
\includegraphics[width=1\columnwidth]{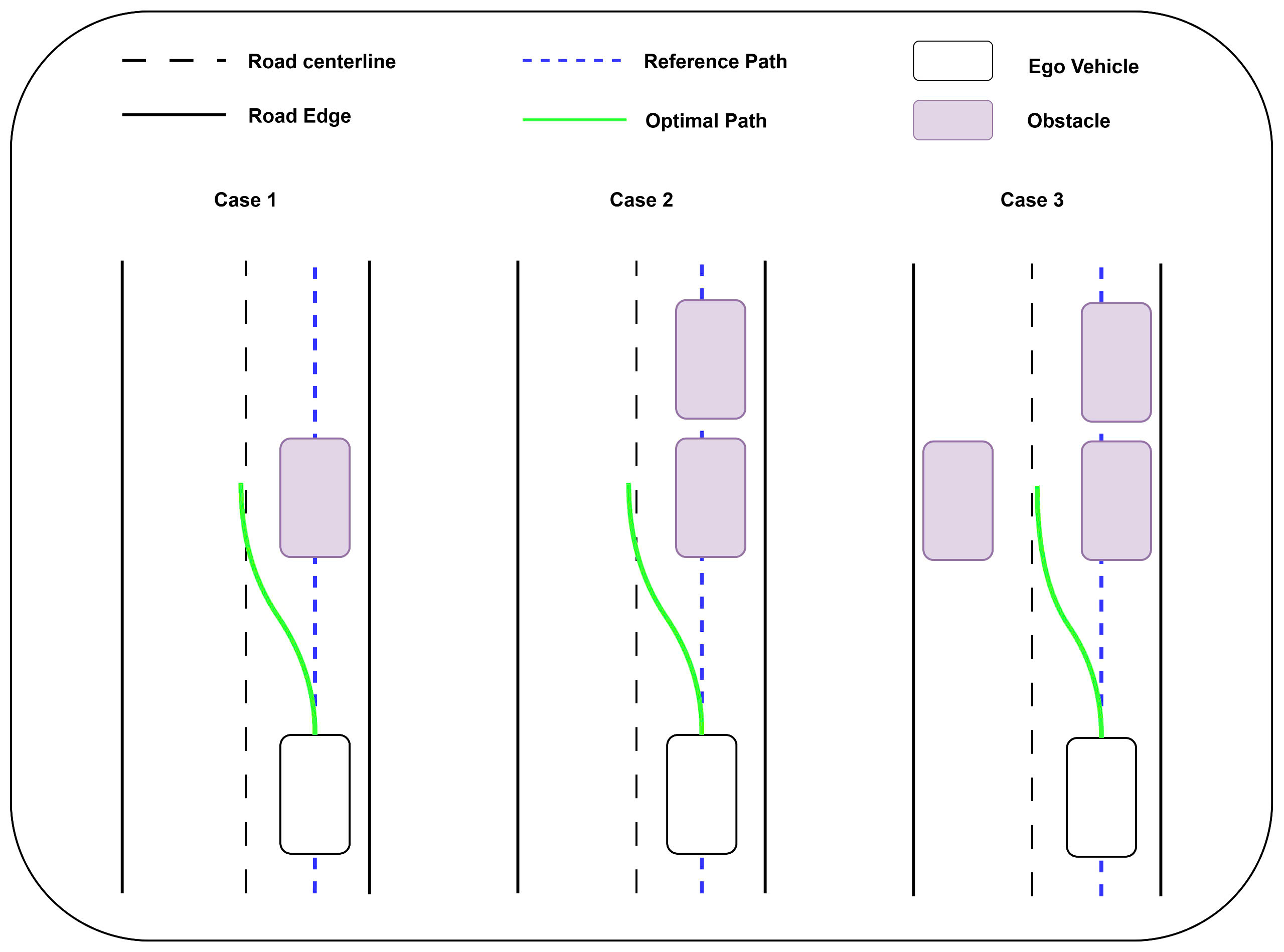}
\caption{Experimental Scenarios for Obstacle Avoidance}
\label{Testing_Scenario}
\end{figure}

\begin{figure}[t]
\centering
\includegraphics[width=1\columnwidth]{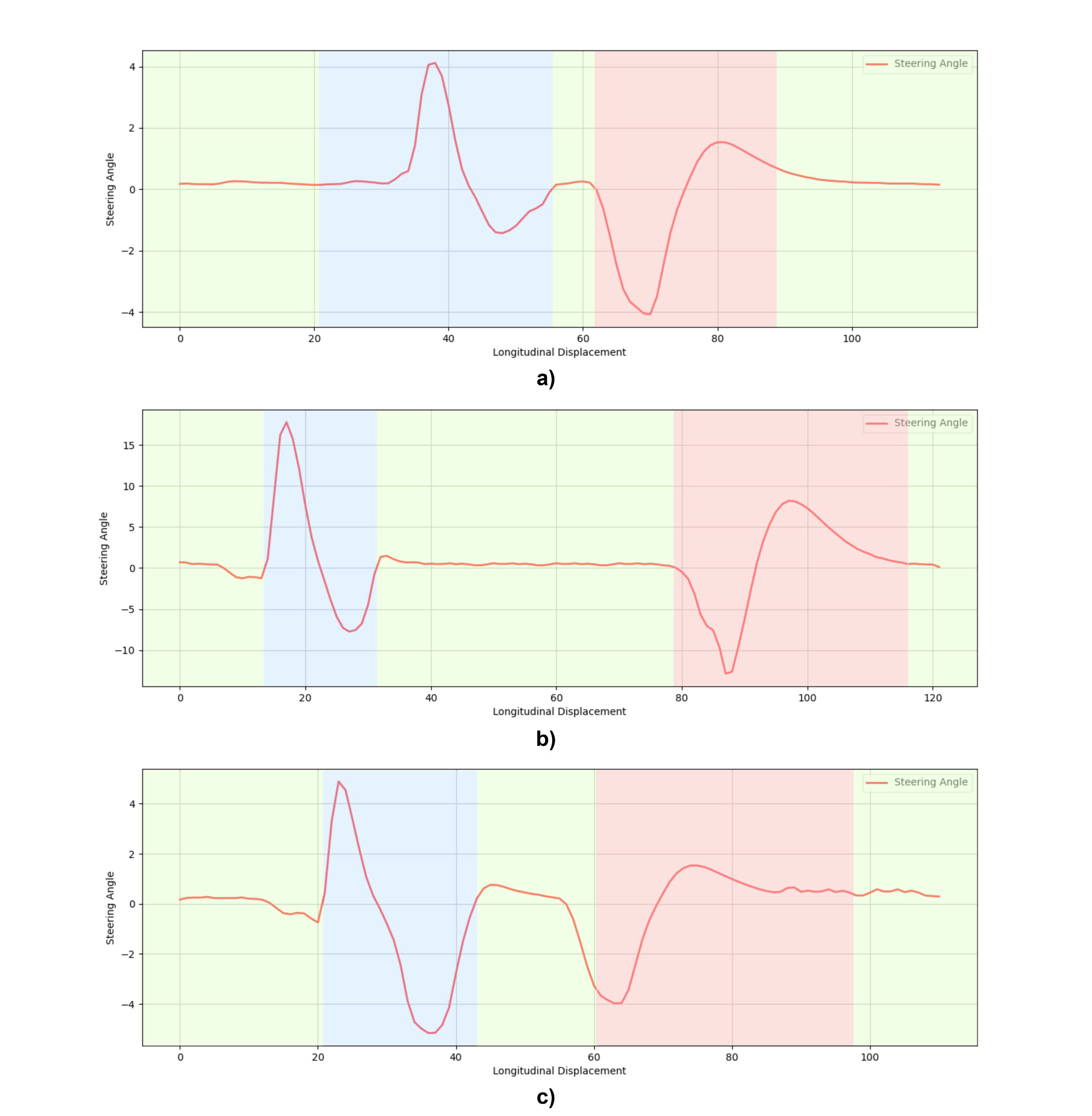}
\caption{Steering Angle of the autonomous vehicle during Obstacle Avoidance. a) case 1, b) case 2, c) case 3. The background color indicates the driving behavior: \textcolor{green}{green} represents straight driving; \textcolor{blue}{blue} represents a left steering maneuver to avoid an obstacle and stabilize the steering angle to keep the vehicle parallel to the obstacle; \textcolor{red}{red} represents a right steering maneuver to return to the original path after bypassing the obstacle and stabilize the steering angle along the reference path.}
\label{fig_steering_angle}
\end{figure}


To evaluate the effectiveness of monocular depth estimation models, we conducted real-world experiments and assessed key performance metrics, including depth accuracy (comprising depth distance and offset distance), as illustrated in Fig. \ref{fig_coordinate_sys_obstacle_avoidance}, and inference time. The evaluation focused on three foundation models: Depth Anything V2 \cite{yang2024depthv2}, MiDaS \cite{Ranftl20MiDaS}, and MonoDepth2 \cite{MonoDepth22019}.

Based on the experimental results in Table \ref{tab:depth_comparison}, Depth Anything V2 \cite{yang2024depthv2} demonstrates the highest accuracy, consistently maintaining low and stable errors across various distances. While MiDaS \cite{Ranftl20MiDaS} achieves slightly better performance at shorter distances, such as 3 meters (0.102 meters error compared to 0.148 meters for Depth Anything V2), its accuracy significantly declines at greater depths. Similarly, Monodepth2 \cite{MonoDepth22019} performs competitively at shorter distances but exhibits substantial errors beyond 8 meters, reaching 3.919 meters at 8 meters and 3.059 meters at 15 meters. MiDaS performs worst at 15 meters, with an absolute error of 5.353 meters, whereas Depth Anything V2 \cite{yang2024depthv2} maintains the lowest error of 0.366 meters, highlighting its superior performance across different depth ranges.

Offset distance also plays a crucial role in ensuring safety during obstacle avoidance. Table \ref{tab:depth_comparison_2} analyzes the offset errors of different models, providing further insights into their reliability Depth Anything V2 \cite{yang2024depthv2} exhibits the most stable and minimal offset across all tested distances, ranging from 0.045 meters to 0.073 meters. In contrast, MiDaS \cite{Ranftl20MiDaS} and Monodepth2 \cite{MonoDepth22019} show greater variations, with Monodepth2 \cite{MonoDepth22019} reaching an offset error of 0.330 meters at 3 meters, indicating less reliable predictions. These results reinforce the robustness of Depth Anything V2 \cite{yang2024depthv2} in terms of both accuracy and consistency.

Table \ref{tab:params_flops_fps} presents the computational performance of the models. Monodepth2 \cite{MonoDepth22019} achieves the highest processing speed at 31 FPS, benefiting from its lightweight architecture (14.33 million parameters and 8.04 GFLOPs). MiDaS \cite{Ranftl20MiDaS}, with its significantly larger computational complexity (344 million parameters and 526.40 GFLOPs), operates at 11 FPS. Depth Anything V2 \cite{yang2024depthv2} strikes a balance, with 24.18 million parameters and 41.30 GFLOPs, achieving a processing speed of 20 FPS. While MiDaS \cite{Ranftl20MiDaS} exhibits the highest computational burden, Depth Anything V2 \cite{yang2024depthv2} offers a better trade-off between speed and efficiency, making it a strong choice for depth estimation tasks that require both accuracy and reasonable inference speed. Therefore, in this study, Depth Anything V2 \cite{yang2024depthv2} is selected to ensure the highest prediction quality and optimal performance.

\subsection{Evaluation of Obstacle Avoidance Performance}




\begin{figure*}[t]
\centering
\includegraphics[width=1\textwidth,height=10cm]{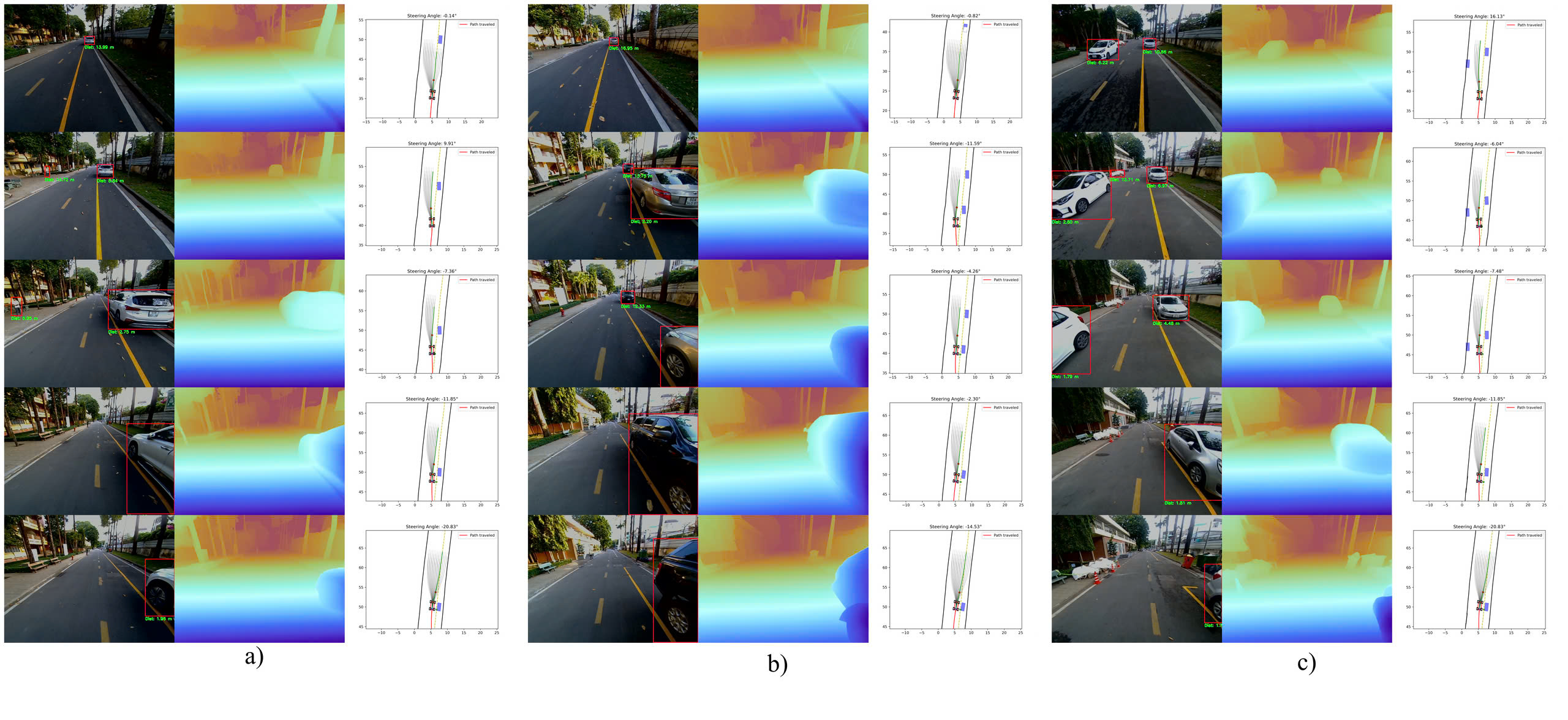}
\caption{Experimental results of obstacle avoidance. (a), (b), and (c) correspond to the cases described in Fig. 4. Each row represents a different timestamp, while the columns visualize the corresponding RGB image with obstacle detection, the depth map estimation, and the real-time simulation visualization.}
\label{fig_exp}
\end{figure*}




To evaluate the obstacle avoidance performance of our algorithm, we conducted experiments under three different scenarios, as described in Fig. \ref{Testing_Scenario}:

- Single obstacle avoidance (case 1): Assessing the vehicle’s ability to detect and navigate around a single stationary object.

- Multiple obstacle avoidance (case 2): Evaluating the system’s performance when encountering multiple static obstacles positioned at various locations.

- Narrow passage navigation (case 3): Testing the capability of the vehicle to maneuver through constrained spaces with obstacles on both sides.

The obstacle avoidance performance is analyzed based on the steering angle of the autonomous vehicle in each test scenario, as shown in Fig. \ref{fig_steering_angle}. The corresponding real-world visualizations for each case are presented in Fig. \ref{fig_exp}. In all three test cases, a consistent pattern comprising five stages is observed. Initially, in the first stage (\textcolor{green}{green phase}), the autonomous vehicle moves straight along the reference path with a near-zero steering angle. Upon detecting an obstacle, it enters the \textcolor{blue}{blue phase}, steering left to avoid the obstacle. It then gradually adjusts the steering angle to negative and subsequently to zero to stabilize on the optimal path. If the obstacle is not yet fully cleared, the vehicle continues moving straight (\textcolor{green}{green phase}). Once the obstacle is completely passed, the vehicle steers right (\textcolor{red}{red phase}) to return to its original trajectory. Before fully stabilizing, it makes a slight left adjustment to refine its position. Finally, as the steering angle returns to zero, the \textcolor{green}{green phase} reappears, indicating that the vehicle is safely following the reference path. Through our experiments, we observed that our algorithmic system is both flexible and stable across various scenarios.

To summarize, the entire system operates at 16 FPS by running depth estimation, detection, and planning modules in parallel. The resulting smooth and safe obstacle avoidance trajectory demonstrates the effectiveness of our proposed algorithm in real-world environments.

\section*{Conclusion}

This paper introduces an efficient perception and planning system for obstacle avoidance, incorporating a camera-based perception module and a motion planning strategy based on Frenet and Pure Pursuit methods. Specifically, the system utilizes YOLOv11 for object detection and the Depth Anything V2 monocular depth estimation model to accurately localize obstacles and enhance situational awareness. The system also evaluates and compares several state-of-the-art YOLO models to identify the most suitable option for obstacle detection. Furthermore, a comparative analysis of advanced monocular depth estimation models under real-world conditions is conducted to examine the trade-offs between accuracy, efficiency, and robustness, providing valuable insights for model selection in practical applications. Experimental results demonstrate the effectiveness of the proposed system in enabling reliable and safe obstacle avoidance.

\textbf{Limitations and future work.} In our experiments, the use of Depth Anything V2 yields acceptable depth estimation accuracy, enabling safe obstacle avoidance within a range of approximately 15 meters. While this distance introduces a practical limitation, it has proven sufficient to ensure safety and stability in our deployment environment. To maintain system simplicity and cost-effectiveness, the perception module relies solely on a monocular camera, avoiding the need for expensive and complex 3D LiDAR sensors. In future work, we aim to enhance the accuracy of depth estimation, optimize real-time performance, and incorporate additional side and rear cameras to address blind spots caused by the single front-facing camera setup, thereby improving the system’s overall perception capabilities.

\section*{Acknowledgment}

This work belongs to the project grant number T2024-02ĐH funded by Ho Chi Minh City University of Technology and Education, Vietnam.

\bibliography{reference} 
\bibliographystyle{ieeetr}
\vspace{12pt}
\color{red}

\end{document}